%% file: rss.tex
\documentclass[conference]{IEEEtran}

\usepackage{times}

\usepackage[numbers]{natbib}
\usepackage{multicol}
\usepackage{dblfloatfix}    

\usepackage{url}
\usepackage{graphicx}
\usepackage{commath}
\usepackage{subcaption}
\usepackage[export]{adjustbox}
\usepackage{wrapfig}
\usepackage[font={small}]{caption}

\usepackage[pagebackref=true,breaklinks=true,letterpaper=true,colorlinks,bookmarks=false]{hyperref}

\usepackage{makerobust}
\usepackage{algorithm}
\usepackage{algpseudocode}
\algnewcommand{\Or}{\ \textbf{or}\ }
\algrenewcommand\algorithmicindent{.5em}%

\MakeRobustCommand{\Call}

\newcommand{\fig}[1]{Fig.~\ref{fig.#1}}
\newcommand{\tabl}[1]{Table~\ref{table.#1}}
\newcommand{\sect}[1]{Sec.~\ref{sec:#1}}

\usepackage{etoolbox}
\newbool{isArxiv}
\booltrue{isArxiv}
\newbool{isIEEE}
\booltrue{isIEEE}

\ifbool{isArxiv}{

    \newcommand{\figdircombined}{combined_rewards}
    \newcommand{\figdirrthreehuman}{r3human}
    \newcommand{\figdirrewards}{rewards}
    \newcommand{\figdirsplits}{splits}
}{ 
    \newcommand{\figdircombined}{combined_rewards}
    \newcommand{\figdirrthreehuman}{r3human}
    \newcommand{\figdirrewards}{rewards}
    \newcommand{\figdirsplits}{splits}
}

\ifbool{isArxiv}{
    \newcommand{\appen}[1]{Appendix~\ref{appendix.#1}}
    \newcommand{\afig}[1]{Fig.~\ref{fig.#1}}
    \newcommand{\afigthree}[3]{Fig. ~\ref{fig.#1}, ~\ref{fig.#2} and ~\ref{fig.#3}}
    \newcommand{\atabl}[1]{Table~\ref{appendix.table.#1}}
}{ 
    \newcommand{\appen}[1]{Appendix~\ref*{appendix.#1} in \citep{SermanetXL16}}
    \newcommand{\afig}[1]{Fig.~\ref*{fig.#1} in \citep{SermanetXL16}}
    \newcommand{\afigthree}[3]{Fig. ~\ref*{fig.#1}, ~\ref*{fig.#2} and ~\ref*{fig.#3} in \citep{SermanetXL16}}
    \newcommand{\atabl}[1]{Table~\ref*{appendix.table.#1} in \citep{SermanetXL16}}
}

\ifbool{isIEEE}{
    \IEEEoverridecommandlockouts
}{ 
    \newcommand{\IEEEauthorrefmark}[1]{}
    \newcommand{\IEEEpeerreviewmaketitle}{}
}

\newcommand{\authorspace}{\hspace{.5cm}}
\graphicspath{ {figures/} }

\title{Unsupervised Perceptual Rewards\\for Imitation Learning}

\author{Pierre Sermanet\IEEEauthorrefmark{1}\thanks{\IEEEauthorrefmark{1} equal contribution} \authorspace Kelvin Xu\IEEEauthorrefmark{1}\IEEEauthorrefmark{2}\thanks{\IEEEauthorrefmark{2} Google Brain Residency program (\href{http://g.co/brainresidency}{g.co/brainresidency})} \authorspace Sergey Levine\\
{\tt\small sermanet,kelvinxx,slevine@google.com}\\
Google Brain\\
}

\begin{document}
\maketitle

\input{rss_body}

\small{
\bibliography{iclr2017_conference}
}

\bibliographystyle{plainnat}

\newpage
\input{rss_appendix}

\end{document}

%% file: rss_body.tex
\begin{abstract}
Reward function design and exploration time are arguably the biggest obstacles to the deployment of reinforcement learning (RL) agents in the real world.
In many real-world tasks, designing a reward function takes considerable hand engineering and often requires additional and potentially visible sensors to be installed just to measure whether the task has been executed successfully.
Furthermore, many interesting tasks consist of multiple implicit intermediate steps that must be executed in sequence. Even when the final outcome can be measured, it does not necessarily provide feedback on these intermediate steps or sub-goals.
To address these issues, we propose leveraging the abstraction power of intermediate visual representations learned by deep models to quickly infer perceptual reward functions from small numbers of demonstrations.
We present a method that is able to identify key intermediate steps of a task from only a handful of demonstration sequences, and automatically identify the most discriminative features for identifying these steps. This method makes use of the features in a pre-trained deep model, but does not require any explicit specification of sub-goals.
The resulting reward functions, which are dense and smooth, can then be used by an RL agent to learn to perform the task in real-world settings.
To evaluate the learned reward functions, we present qualitative results on two real-world tasks and a quantitative evaluation against a human-designed reward function. We also demonstrate that our method can be used to learn a complex real-world door opening skill using a real robot, even when the demonstration used for reward learning is provided by a human using their own hand.
To our knowledge, these are the first results showing that complex robotic manipulation skills can be learned directly and without supervised labels from a video of a human performing the task. Supplementary material and dataset are available at \href{https://sermanet.github.io/rewards/}{sermanet.github.io/rewards}

\end{abstract}

\IEEEpeerreviewmaketitle

\section{Introduction}

Social learning, such as imitation, plays a critical role in allowing humans and animals to acquire complex skills in the real world. Humans can use this weak form of supervision to acquire behaviors from very small numbers of demonstrations, in sharp contrast to deep reinforcement learning (RL) methods, which typically require extensive training data. In this work, we make use of two ideas to develop a scalable and efficient imitation learning method: first, imitation makes use of extensive prior knowledge to quickly glean the ``gist'' of a new task from even a small number of demonstrations; second, imitation involves both observation and trial-and-error learning (RL). Building on these ideas, we propose a reward learning method for understanding the intent of a user demonstration through the use of pre-trained visual features, which provide the ``prior knowledge'' for efficient imitation. Our algorithm aims to discover not only the high-level goal of a task, but also the implicit sub-goals and steps that comprise more complex behaviors. Extracting such sub-goals can allow the agent to make maximal use of information contained in a demonstration. Once the reward function has been extracted, the agent can use its own experience at the task to determine the physical structure of the behavior, even when the reward is provided by an agent with a substantially different embodiment (e.g. a human providing a demonstration for a robot).

\begin{figure}
\begin{center}
  \includegraphics[width=1\linewidth]{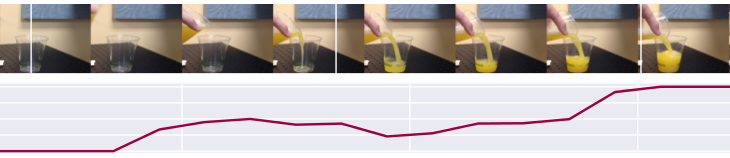}
  \includegraphics[width=1\linewidth]{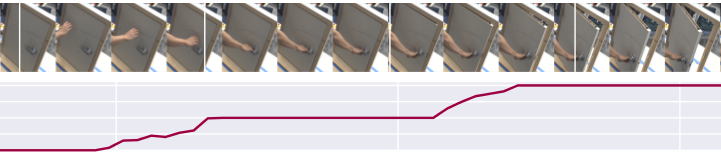}
  \includegraphics[width=1\linewidth]{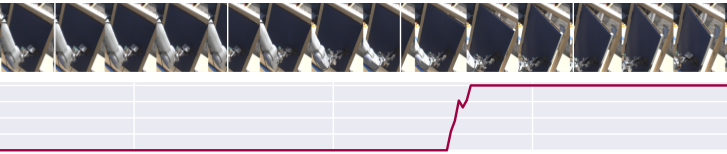}
\end{center}
\caption{{\bf Perceptual reward functions learned from unsupervised observation of human demonstrations.} For each video strip, we show its corresponding learned reward function below with the range $[0, 4]$ on the vertical axis, where $4$ corresponds to the maximum reward for completing the demonstrated task. We show rewards learned from human demonstrations for a pouring task (top) and door opening task (middle) and use the door opening reward to successfully train a robot to perform the task (bottom).
}
\label{fig.r3human.rewards_summary}
\end{figure}

\begin{figure*}[ht]
\begin{center}
\includegraphics[width=.9\linewidth]{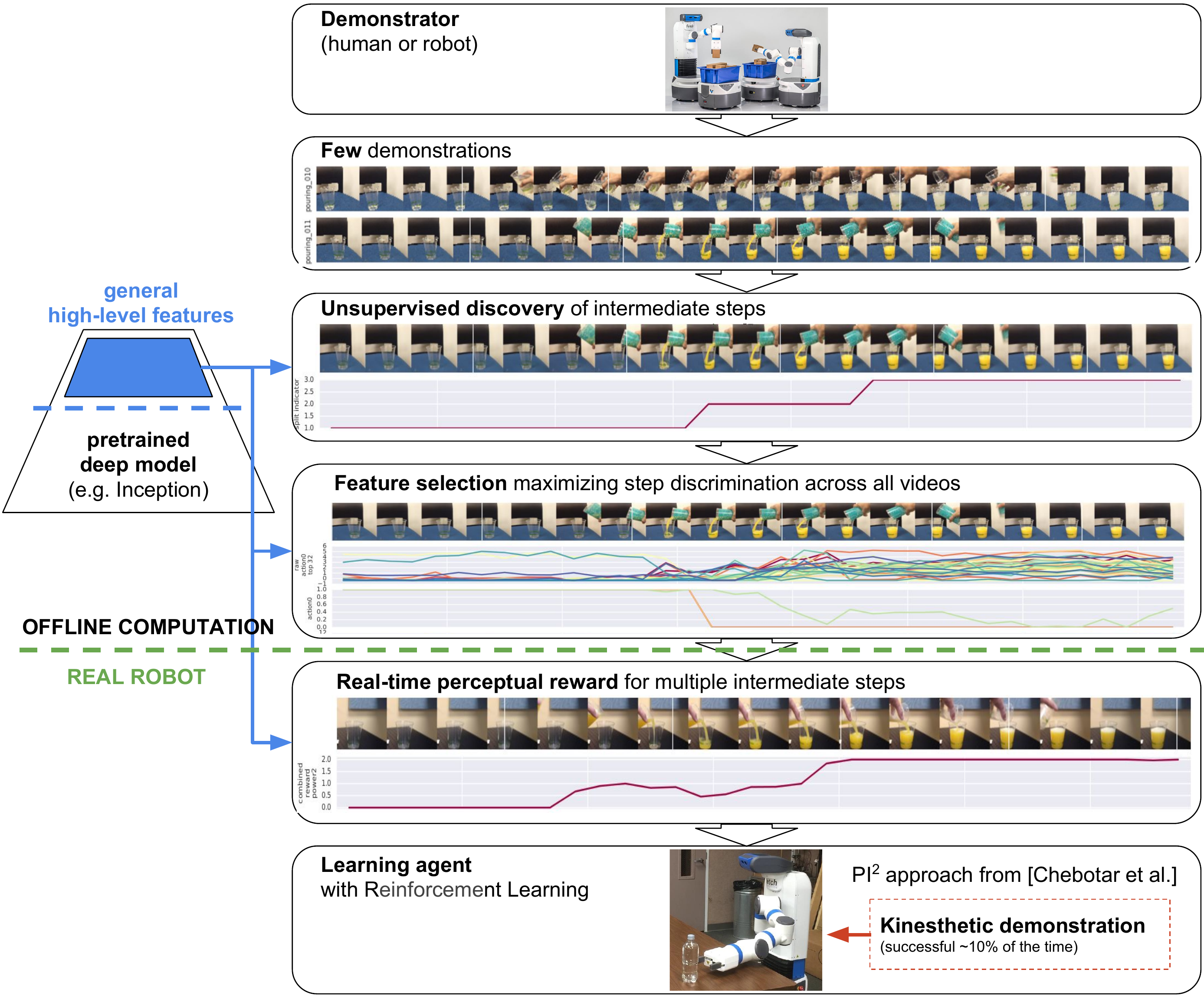}
\end{center}
\caption{{\bf Method overview.} Given a few demonstration videos of the same action, our method discovers intermediate steps,
then trains a classifier for each step on top of the mid and high-level representations of a pre-trained deep model (in this work, we use all activations starting from the first ``mixed'' layer that follows the first 5 convolutional layers). The step classifiers are then combined to produce a single reward function per step prior to learning.
These intermediate rewards are combined into a single reward function.
The reward function is then used by a real robot to learn the perform the demonstrated task as shown in \sect{door}.
}
\label{fig.overview}
\end{figure*}

To our knowledge, our method is the first reward learning technique that learns generalizable vision-based reward functions for complex robotic manipulation skills from only a few demonstrations provided directly by a human. Although prior methods have demonstrated reward learning with vision for real-world robotic tasks, they have either required kinesthetic demonstrations with robot state for reward learning \citep{finn2015deep}, or else required low-dimensional state spaces and numerous demonstrations \citep{WulfmeierIROS2016}. The contributions of this paper are:
\begin{itemize}
\item A method for perceptual reward learning from only a {\bf few demonstrations} of {\bf real-world} tasks. Reward functions are dense and incremental, with automated {\bf unsupervised} discovery of intermediate steps.
\item {\bf The first vision-based reward learning method} that can learn a complex robotic manipulation task from a few human demonstrations in {\bf real-world robotic experiments}.
\item A set of empirical experiments that show that the learned visual representations inside a pre-trained deep model are general enough to be directly used to represent goals and sub-goals for manipulation skills in new scenes {\bf without retraining}.
\end{itemize}

\subsection{Related Work}

Our method can be viewed as an instance of learning from demonstration (LfD), which has long been studied as a strategy to overcome the challenging problem of learning robotic tasks from scratch \citep{calinon2008} (also see \citep{argall2009survey} for a survey). Previous work in LfD has successfully replicated gestures \cite{calinon2007incremental} and dual-arm manipulation \citep{asfour2008imitation} on real robotic systems. LfD can however be challenging when the human demonstrator and robot do not share the same ``embodiment''. A mismatch between the robot's sensor/actions and the recorded demonstration can lead to suboptimal behaviour. Previous work has addressed this issue by either providing additional feedback \citep{wrede2013user}, or alternatively providing demonstrations which only give a coarse solutions which can be iteratively refined \citep{prieur2012modeling}. Our method of using perceptual rewards can be seen as an attempt to circumvent the need for shared embodiment in the general problem of learning from humans. To accomplish this, we first learn a visual reward from human demonstrations, and then learn \emph{how} to perform the task using the robot's own experience via reinforcement learning. 

Our approach can be seen as an instance of the more general inverse reinforcement learning framework \citep{ng2000algorithms}. Inverse reinforcement learning can be performed with a variety of algorithms, ranging from margin-based methods \citep{abbeel2004apprenticeship,ratliff2006maximum} to methods based on probabilistic models \citep{ramachandran2007bayesian,ziebart2008maximum}. Most similar in spirit to our method is the recently proposed SWIRL algorithm \citep{krishnan2016swirl} and non-parametric reward segmentation approaches \citep{ranchod2015,michini2013,niekum2015}, which like our method attempts to decompose the task into sub-tasks to learn a sequence of local rewards. These approaches differs in that they consider only problems with low-dimensional state spaces while our method can be applied to raw pixels. Our experimental evaluation shows that our method can learn a complex real-world door opening skill using visual rewards from a human demonstration. 

We use visual features obtained from training a deep convolutional network on a standard image classification task to represent our visual rewards. Deep learning has been used in a range of applications within robotic perception and learning \citep{sung2015robobarista,pinto2016supersizing,byravan2016se3}. In work that has specifically examined the problem of learning reward functions from images, one common approach to image-based reward functions has been to directly specify a ``target image'' by showing the learner the raw pixels of a successful task completion state, and then using distance to that image (or its latent representation) as a reward function \citep{lange2012autonomous,finn2015deep,watter2015embed}.
However, this approach has several shortcomings. First, the use of a target image presupposes that the system can achieve a substantially similar visual state, which precludes generalization to semantically similar but visually distinct situations. Second, the use of a target image does not provide the learner with information about which facet of the image is more or less important for task success, which might result in the learner excessively emphasizing irrelevant factors of variation (such as the color of a door due to light and shadow) at the expense of relevant factors (such as whether or not the door is open or closed).

A few recently proposed IRL algorithms have sought to combine IRL with vision and deep network representations \citep{finn2016guided,WulfmeierIROS2016}.
However, scaling IRL to high-dimensional systems and open-ended reward representations is very challenging. The previous work closest to ours used images together with robot state information (joint angles and end effector pose), with demonstrations provided through kinesthetic teaching \citep{finn2016guided}. The approach we propose in this work, which can be interpreted as a simple and efficient approximation to IRL, can use demonstrations that consist of videos of a human performing the task using their own body, and can acquire reward functions with intermediate sub-goals using just a few examples. This kind of efficient vision-based reward learning from videos of humans has not been demonstrated in prior IRL work. The idea of perceptual reward functions using raw pixels was also explored by \cite{edwards2016perceptual} which, while sharing the same spirit as this work, was limited to synthetic tasks and used single images as perceptual goals rather than multiple demonstration videos.

\section{Simple Inverse Reinforcement Learning with Visual Features}

The key insight in our approach is that we can exploit the semantically meaningful and powerful features in a pre-trained deep neural network to infer task goals and sub-goals using a very simple approximate inverse reinforcement learning method. The pre-trained network effectively transfers prior knowledge about the visual world to make imitation learning fast and robust. Our approach can be interpreted as a simple approximation to inverse reinforcement learning under a particular choice of system dynamics, as discussed in Section~\ref{sec:irl}. While this approximation is somewhat simplistic, it affords an efficient and scalable learning rule that minimizes overfitting even when trained on a small number of demonstrations. As depicted in \fig{overview}, our algorithm first segments the demonstrations into segments based on perceptual similarity, as described in Section~\ref{sec:seg}. Intuitively, the resulting segments correspond to sub-goals or steps of the task. The segments can then be used as a supervision signal for learning steps classifiers, described in Section~\ref{sec:selection}, which produces a single perception reward function for each step of the task. The combined reward function can then be used with a reinforcement learning algorithm to learn the demonstrated behavior. Although this method for extracting reward functions is exceedingly simple, its power comes from the use of highly general and robust pre-trained visual features, and our key empirical result is that such features are sufficient to acquire effective and generalizable reward functions for real-world manipulation skills.

We use the Inception network \citep{DBLP:journals/corr/SzegedyVISW15} pre-trained for ImageNet classification \citep{imagenet_cvpr09} to obtain the visual features for representing the learned rewards. It is well known that visual features in such networks are quite general and can be reused for other visual tasks. However, it is less clear if sparse subsets of such features can be used directly to represent goals and sub-goals for real-world manipulation skills. Our experimental evaluation suggests that indeed they can, and that the resulting reward representations are robust and reliable enough for real-world robotic learning without any finetuning of the features. In this work, we use all activations starting from the first ``mixed'' layer that follows the first 5 convolutional layers (this layer's activation map is of size 35x35x256 given a 299x299 input). While this paper focuses on visual perception, the approach could be generalized to other modalities (e.g. audio and tactile).

\subsection{Inverse Reinforcement Learning with Time-Independent Gaussian Models}
\label{sec:irl}
In this work, we use a very simple approximation to the MaxEnt IRL model~\citep{ziebart2008maximum}, a popular probabilistic approach to IRL. We will use $s_t$ to denote the visual feature activations at time $t$, which constitute the state, $s_{it}$ to denote the $i^\text{th}$ feature at time $t$, and $\tau = \{s_1,\dots,s_T\}$ to denote a sequence or trajectory of these activations in a video. In MaxEnt IRL, the demonstrated trajectories $\tau$ are assumed to be drawn from a Boltzmann distribution according to:
\begin{equation}
p(\tau) = p(s_1,\dots,s_T) = \frac{1}{Z}\exp\left( \sum_{t=1}^T R(s_t) \right),\label{eqn:maxent}
\end{equation}
\noindent where $R(s_t)$ is the unknown reward function. The principal computational challenge in MaxEnt IRL is to approximate $Z$, since the states at each time step are not independent, but are constrained by the system dynamics. In deterministic systems, where $s_{t+1} = f(s_t, a_t)$ for actions $a_t$ and dynamics $f$, the dynamics impose constraints on which trajectories $\tau$ are feasible. In stochastic systems, where $s_{t+1} \sim p(s_{t+1} | s_t, a_t)$, we must also account for the dynamics distribution in Equation~(\ref{eqn:maxent}), as discussed by \cite{ziebart2008maximum}.
Prior work has addressed this using dynamic programming to exactly compute $Z$ in small, discrete systems~\citep{ziebart2008maximum}, or by using sampling to estimate $Z$ for large, continuous domains~\citep{kalakrishnan2010inverse,boularias2011relative,finn2016guided}.
Since our state representation corresponds to a large vector of visual features, exact dynamic programming is infeasible. Sample-based approximation requires running a large number of trials to estimate $Z$ and, as shown in recent work~\citep{finn2016connection},
corresponds to a variant of generative adversarial networks (GANs), with all of the accompanying stability and optimization challenges. Furthermore, the corresponding model for the reward function is complex, making it prone to overfitting when only a small number of demonstrations is available.

When faced with a difficult learning problem in extremely low-data regimes, a standard solution is to resort to simple, biased models, so as to minimize overfitting. We adopt precisely this approach in our work: instead of approximating the complex posterior distribution over trajectories under nonlinear dynamics, we use a simple biased model that affords efficient learning and minimizes overfitting. Specifically, we assume that all trajectories are dynamically feasible, and that the distribution over each activation at each time step is independent of all other activations and all other time steps. This corresponds to the IRL equivalent of a na\"{i}ve Bayes model: in the same way that na\"{i}ve Bayes uses an independence assumption to mitigate overfitting in high-dimensional feature spaces, we use independence between both time steps and features to learn from very small numbers of demonstrations. Under this assumption, the probability of a trajectory $\tau$ factorizes according to
\[
p(\tau) = \prod_{t=1}^T \prod_{i=1}^N p(s_{it}) = \prod_{t=1}^T \prod_{i=1}^N \frac{1}{Z_{it}} \exp\left( R_{i}(s_{it}) \right),
\]
\noindent which corresponds to a reward function of the form $R_t(s_t) = \sum_{i=1}^N R_{i}(s_{it})$. We can then simply choose a form for $R_{i}(s_{it})$ that can be normalized analytically, such as one which is quadratic in $s_{it}$ which yields a Gaussian distribution. While this approximation is quite drastic, it yields an exceedingly simple learning rule: in the most basic version, we have only to fit the mean and variance of each feature distribution, and then use the log of the resulting Gaussian as the reward.

\begin{algorithm}
  \begin{algorithmic}
\small{
\Function{Split}{$video, start, end, n, min\_size, prev\_std=[]$}
  \If{$n = 1$}
  \Return $[], [\Call{AverageStd}{video[start:end]}]$
  \EndIf
  \State $min\_std \leftarrow None$
  \State $min\_std\_list \leftarrow []$
  \State $min\_split \leftarrow []$
  \For{$i \leftarrow start + min\_size$ \textbf{to} $end - ((n - 1) * min\_size))$}
    \State $std1 \leftarrow [\Call{AverageStd}{video[start:i]}]$
    \State $splits2, std2 \leftarrow \textsc{Split}(\parbox[t]{.5\linewidth}{$video$, $i$, $end$, $n - 1$, $min\_size$, $std1 + prev\_std$)}$
    \State $avg\_std \leftarrow \Call{AverageStd}{\Call{Join}{prev\_std, std1, std2}}$
    \If {$min\_std = None \Or avg\_std < min\_std$}
      \State $min\_std \leftarrow avg\_std$
      \State $min\_std\_list \leftarrow \Call{Join}{std1, std2}$
      \State $min\_split \leftarrow \Call{Join}{i, splits2}$
    \EndIf
  \EndFor
  \State \Return $min\_split, min\_std\_list$
\EndFunction
\caption{Recursive similarity maximization, where
$AverageStd()$ is a function that computes the average standard deviation over a set of frames or over a set of values, $Join()$ is a function that joins values or lists together into a single list, $n$ is the number of splits desired and $min\_size$ is the minimum size of a split.}
\label{algo.similarity}
}
\end{algorithmic}
\end{algorithm}

\subsection{Discovery of Intermediate Steps}
\label{sec:seg}

The simple IRL model in the previous section can be used to acquire a single quadratic reward function in terms of the visual features $s_t$. However, for complex multi-stage tasks, this model can be too coarse, making task learning slow and difficult. We therefore instead fit multiple quadratic reward functions, with one reward function per intermediate step or goal. These steps are discovered automatically in the first stage of our method, which is performed independently on each demonstration. If multiple demonstrations are available, they are pooled together in the feature selection step discussed in the next section, and could in principle be combined at the segmentation stage as well, though we found this to be unnecessary in our prototype. The intermediate steps model extends the simple independent Gaussian model in the previous section by assuming that
\[
p(\tau) = \prod_{t=1}^T \prod_{i=1}^N \frac{1}{Z_{it}} \exp\left( R_{i{g_t}}(s_{it}) \right),
\]
where $g_t$ is the index of the goal or step corresponding to time step $t$. Learning then corresponds to identifying the boundaries of the steps in the demonstration, and fitting independent Gaussian feature distributions at each step. Note that this corresponds exactly to segmenting the demonstration such that the variance of each feature within each segment is minimized.

In this work, we employ a simple recursive video segmentation algorithm
as described in Algorithm~\ref{algo.similarity}. Intuitively, this method breaks down a sequence in a way that each frame in a segment is abstractly similar to each other frame in that segment. The number of segments is treated like a hyperparameter in this approach. There exists a body of 
unsupervised video segmentation methods \cite{yuan2007formal,kroemer2014} which
would likely enable a less constrained set of demonstrations to be used. While this is an important avenue of future work, we show that
our simple approach is sufficient to demonstrate the efficacy of our method on a realistic set of demonstrations. We also investigate how to reduce the search space of similar feature patterns across videos in section~\ref{sec:selection}. This would render discovery of video alignments tractable for an optimization method such as the one used in~\cite{tangECCV14} for video co-localization. 

The complexity of Algorithm~\ref{algo.similarity} is $\mathcal{O}(n^{m})$ where $n$ is the number of frames in a sequence and $m$ the number of splits. We also experiment with a greedy binary version of this algorithm (Algorithm~\ref*{algo.binary_similarity} detailed in \appen{binary_similarity}): first split the entire sequence in two, then recursively split each new segment in two. While not exactly minimizing the variance across all segments, it is significantly more efficient ($\mathcal{O}(n^2\log{}m)$) and yields qualitatively sensible results.

\subsection{Steps Classification}
\label{sec:selection}

In this section we explore learning a steps classifier on top of the pre-trained deep model using a regular linear classifier and a custom feature selection classifier.

\begin{figure}[h]
\begin{center}
\includegraphics[width=1\linewidth]{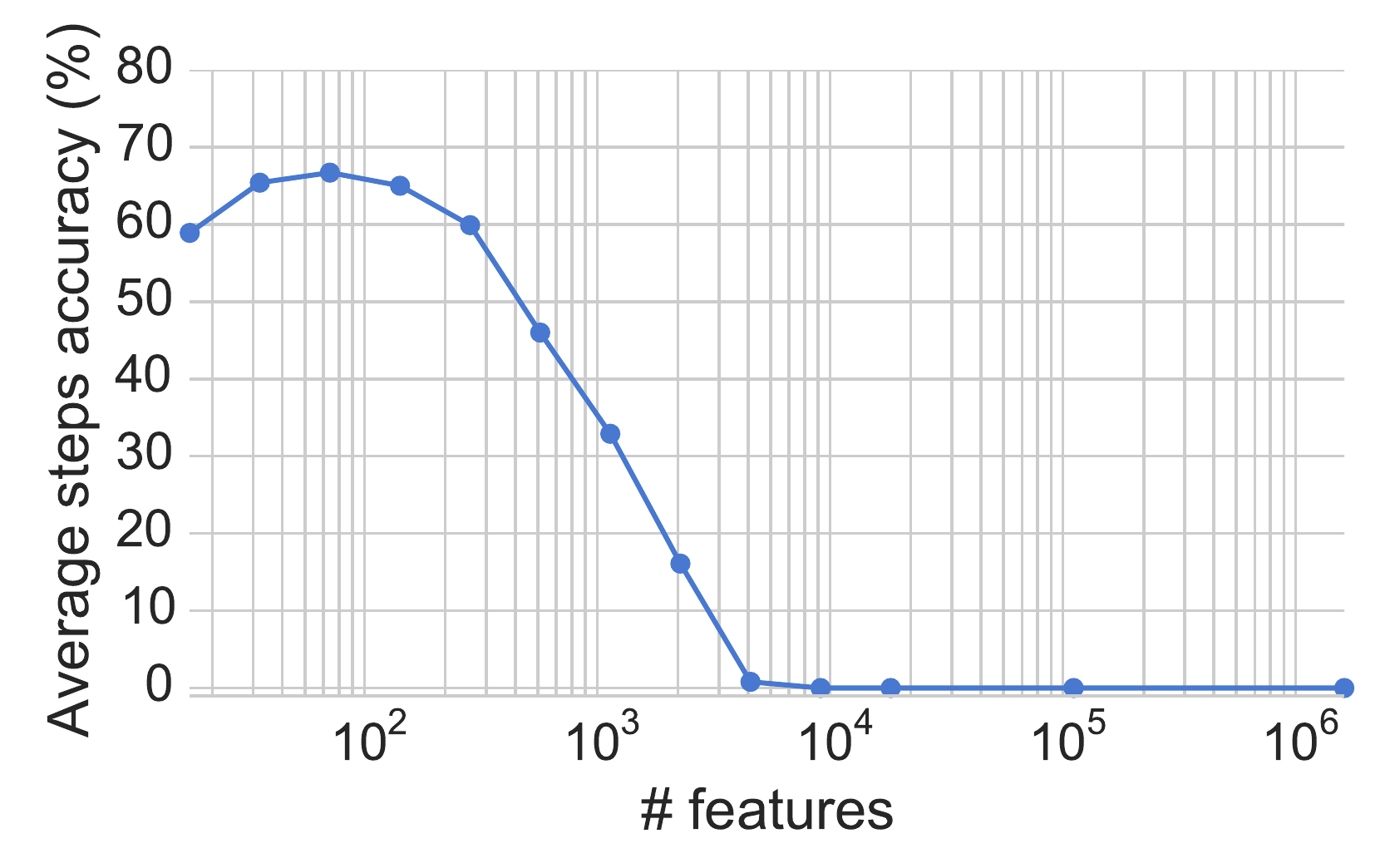}
\end{center}
\caption{{\bf Feature selection classification accuracy} on the pouring validation set for 2-steps classification. By varying the number of features $n$ selected, we show that the method yields good results in the region $n = [32, 64]$, but collapses to 0\% accuracy starting at $n = 8192$.}
\label{fig.features_graph}
\end{figure}

One could consider a naive Bayes classifier here, however we show in \fig{features_graph} that it overfits badly when using all features. Other options are to select a small set of features, or use an alternative classification approach. We first discuss a feature selection approach that is efficient and simple. Then we discuss the alternative logistic regression approach which is simple but less principled. As our experimental results in \tabl{rewards_accuracy} show, the performance slightly exceeds the selection method, though requires using all of the features.

Intent understanding requires identifying highly discriminative features of a specific goal while remaining invariant to unrelated variation (e.g. lighting, color, viewpoint). The relevant discriminative features may be very diverse and more or less abstract, which motivates our intuition to tap into the activations of deep models at different depths. Deep models cover a large set of representations that can be useful, from spatially dense and simple features in the lower layers (e.g. large collection of detected edges) to gradually more spatially sparse and abstract features (e.g. few object classes).

Within this high-dimensional representation, we hypothesize that there exists a subset of mid to high-level features that can readily and compactly discriminate previously unseen inputs. We investigate that hypothesis using a simple feature selection method described in \appen{selection}. The existence of a small subset of discriminative features can be useful for reducing overfitting in low data regimes, but more importantly can allow drastic reduction of the search space for the unsupervised steps discovery. Indeed since each frame is described by millions of features, finding patterns of feature correlations across videos leads to a combinatorial explosion. However, the problem may become tractable if there exists a low-dimensional subset of features that leads to reasonably accurate steps classification. We test and discuss that hypothesis in Section~\ref{sec:quantitative_analysis}.

We also train a simple linear layer which takes as input the same mid to high level activations used for steps discovery and outputs a score for each step. This linear layer is trained using logistic regression and the underlying deep model is not fine-tuned. Despite the large input (1,453,824 units) and the low data regime (11 to 19 videos of ~30 to 50 frames each), we show that this model does not severely overfit to the training data and perform slightly better than the feature selection method as described in Section~\ref{sec:quantitative_analysis}.

\subsection{Using Perceptual Rewards for Robotic Learning}
\label{sec:rl}

In order to use our learned perceptual reward functions in a complete skill learning system, we must also choose a reinforcement learning algorithm and a policy representation. While in principle any reinforcement learning algorithm could be suitable for this task, we chose a method that is efficient for evaluation on real-world robotic systems in order to validate our approach. The method we use is based on the PI$^2$ reinforcement learning algorithm \citep{theodorou2010generalized}. Our implementation, which is discussed in more detail in \appen{rl_details}, uses a relatively simple linear-Gaussian parameterization of the policy, which corresponds to a sequence of open-loop torque commands with fixed linear feedback to correct for perturbations. This method also requires initialization from example demonstrations to learn complex manipulation tasks efficiently. A more complex neural network policy could also be used~\citep{chebotar2016path}, and more sophisticated RL algorithms could also learn skills without demonstration initialization. However, since the main purpose of this component is to validate the learned reward functions, we used this simple approach to test our rewards quickly and efficiently.

\section{Experiments}

In this section, we discuss our empirical evaluation, starting with an analysis of the learned reward functions in terms of both qualitative reward structure and quantitative segmentation accuracy. We then present results for a real-world validation of our method on robotic door opening.

\subsection{Perceptual Rewards Evaluation}

\begin{figure}[h]
\begin{center}
\includegraphics[width=1\linewidth]{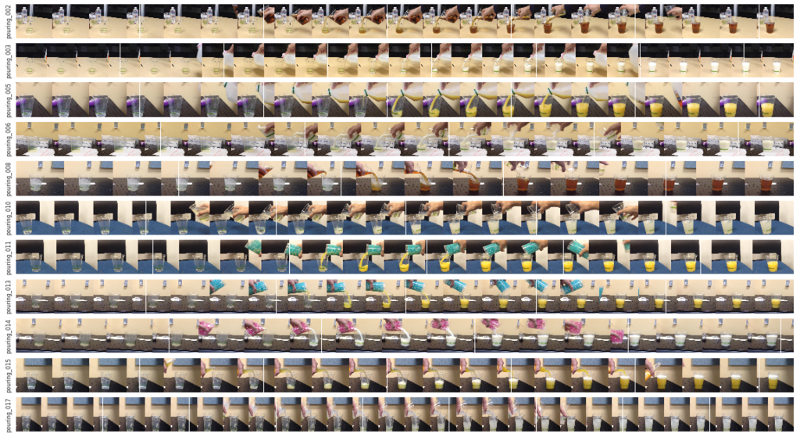}
\end{center}
\caption{{\bf Entire training set for the pouring task} (11 demonstrations).}
\label{fig.data.pouring}
\end{figure}

We report results on two demonstrated tasks: door opening and liquid pouring. We collected about a dozen training videos for each task using a smart phone. As an example, \fig{data.pouring} shows the entire training set used for the pouring task.

\subsubsection{Qualitative Analysis}

\begin{figure*}[htb]
\begin{center}

\adjustbox{minipage=1.3em,valign=t}{\subcaption{}\label{fig.pouring_016_short}}
\begin{subfigure}[t]{.9\linewidth}
  \includegraphics[width=1\linewidth]{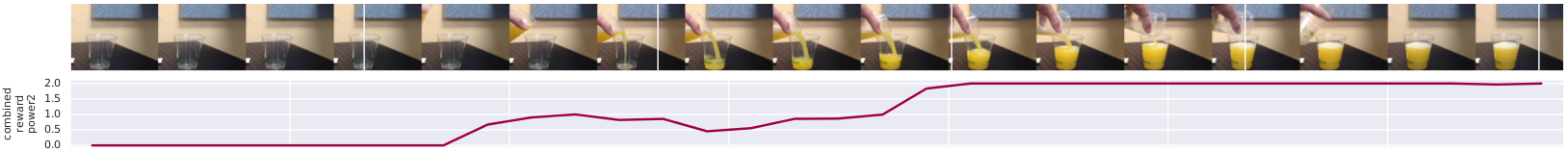}
\end{subfigure}

\adjustbox{minipage=1.3em,valign=t}{\subcaption{}\label{fig.pouring-miss_001_short}}
\begin{subfigure}[t]{.9\linewidth}
  \includegraphics[width=1\linewidth]{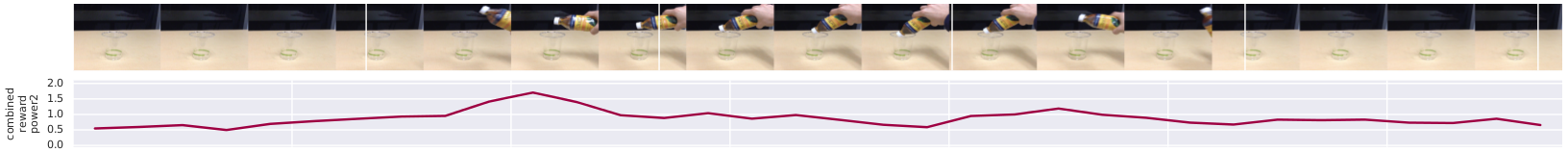}
\end{subfigure}

\adjustbox{minipage=1.3em,valign=t}{\subcaption{}\label{fig.pouring-miss_004_short}}
\begin{subfigure}[t]{.9\linewidth}
  \includegraphics[width=1\linewidth]{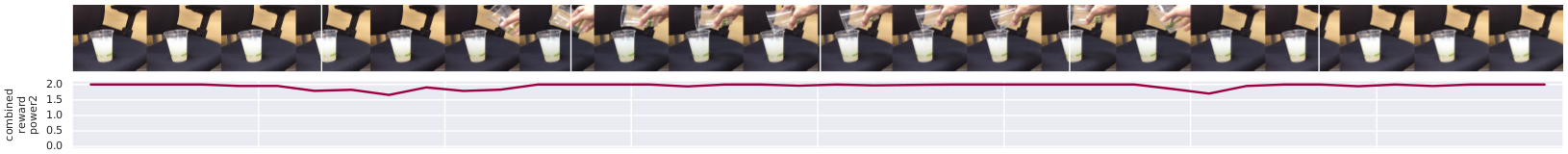}
\end{subfigure}

\end{center}
\caption{{\bf Examples of "pouring" reward functions.} We show here a few successful examples, see \afig{qual.rewards} for results on the entire test set. In \fig{pouring_016_short} we observe a continuous and incremental reward as the task progresses and saturating as it is completed. \fig{pouring-miss_001_short} increases as the bottle appears but successfully detects that the task is not completed, while in \fig{pouring-miss_004_short} it successfully detects that the action is already completed from the start.}
\label{fig.qual.rewards.short}
\end{figure*}

While a door opening sensor can be engineered using sensors hidden in the door, measuring pouring or container tilting would be quite complicated, would visually alter the scene, and is unrealistic for learning in the wild. Visual reward functions are therefore an excellent choice for complex physical phenomena such as liquid pouring. 
In \fig{qual.rewards.short}, we present the combined reward functions for test videos on the pouring task, and \afig{qual.door} shows the intermediate rewards for each sub-goal. We plot the predicted reward functions for both successful and failed task executions in \afig{qual.rewards}. We observe that for ``missed" executions where the task is only partially performed, the intermediate steps are correctly classified.
\afig{splits} details qualitative results of unsupervised step segmentation for the door opening and pouring tasks. For the door task, the 2-segments splits are often quite in line with what one can expect, while a 3-segments split is less accurate. We also observe that the method is robust to the presence or absence of the handle on the door, as well as its opening direction. We find that for the pouring task, the 4-segments split often yields the most sensible break down. It is interesting to note that the 2-segment split usually occurs when the glass is about half full.

{\bf Failure Cases}

The intermediate reward function for the door opening task which corresponds to a human hand manipulating the door handle seems rather noisy or wrong in \afigthree{X_11_1}{hobbit_7}{1965_0} ("action1" on the y-axis of the plots).The reward function in \afig{pouring_009} remains flat while liquid is being poured into the glass. The liquid being somewhat transparent, we suspect that it looks too similar to the transparent glass for the function to fire.

\subsubsection{Quantitative Analysis}
\label{sec:quantitative_analysis}

We evaluate the quantitative accuracy of the unsupervised steps discovery in \tabl{split_accuracy} (see \atabl{split_accuracy} for more details), while \tabl{rewards_accuracy} presents quantitative generalization results for the learned reward on a test video of each task. For each video, ground truth intermediate steps were provided by human supervision for the purpose of evaluation. While this ground truth is subjective, since each task can be broken down in multiple ways, it is consistent for the simple tasks in our experiments. We use the Jaccard similarity measure (intersection over union) to indicate how much a detected step overlaps with its corresponding ground truth.

\begin{table}[h]
\small
\caption{{\bf Unsupervised steps discovery accuracy} (Jaccard overlap on training sets) versus the ordered random steps baseline.}\label{table.split_accuracy}
\begin{center}

\begin{tabular}{c|c|c|c}
{\bf dataset} & {\bf method} & {\bf 2 steps}  & {\bf 3 steps}\\
{\bf (training)} & & {\bf average} & {\bf average}\\
\hline
door & ordered random steps & 52.5\% &  55.4\% \\
     & unsupervised steps   & \bf 76.1\% & \bf 66.9\% \\
 \hline
pouring & ordered random steps & 65.9\% & 52.9\% \\
        & unsupervised steps   & \bf 91.6\% & \bf 58.8\% \\
\end{tabular}
\end{center}
\end{table}

In \tabl{split_accuracy}, we compare our method against a random baseline. Because we assume the same step order in all demonstrations, we also order the random steps in time to provide a fair baseline. Note that the random baseline performs fairly well because the steps are distributed somewhat uniformly in time. Should the steps be much less temporally uniform, the random baseline would be expected to perform very poorly, while our method should maintain similar performance. We compare splitting between 2 and 3 steps and find that, for both tasks, 2 steps are easier to discover, probably because these tasks exhibit one strong visual change each while the other steps are more subtle.
Note that our unsupervised segmentation only works when full sequences are available while our learned reward functions can be used in real-time without accessing future frames. Hence in these experiments we evaluate the unsupervised segmentation on the training set only and evaluate the reward functions on the test set.

\begin{table}[h]
\small
\caption{{\bf Reward functions accuracy} by steps (Jaccard overlap on test sets).}
\label{table.rewards_accuracy}
\begin{center}
\begin{tabular}{c|c|c|c}
{\bf dataset} & {\bf classification} & {\bf 2 steps}  & {\bf 3 steps}\\
{\bf (testing)} & {\bf method} & {\bf average} & {\bf average}\\
\hline
door & random baseline          & $33.6\% \pm 1.6$ & $25.5\% \pm 1.6$ \\
     & feature selection & $ 72.4\% \pm 0.0$ & $ 52.9\% \pm 0.0$ \\
     & linear classifier & ${\bf 75.0\% \pm 5.5} $ & ${\bf 53.6\% \pm 4.7 }$ \\
 \hline
pouring & random baseline & $31.1\% \pm 3.4 $ & $ 25.1\% \pm 0.1 $\\
        & feature selection & $65.4\% \pm 0.0$ & $40.0\% \pm 0.0$ \\
        & linear classifier & $ {\bf 69.2\% \pm 2.0}$ & ${\bf 49.6\% \pm 8.0}$\\
\end{tabular}
\end{center}
\end{table}

In \tabl{rewards_accuracy}, we evaluate the reward functions individually for each step on the test set. For that purpose, we binarize the reward function using a threshold of $0.5$. The random baseline simply outputs true or false at each timestep. We observe that the learned feature selection and linear classifier functions outperform the baseline by about a factor of $2$. It is not clear exactly what the minimum level of accuracy is required to successfully learn to perform these tasks, but we show in section~\ref{sec.real.qualitative} that the reward accuracy on the door task is sufficient to reach $100\%$ success rate with a real robot. Individual steps accuracy details can be found in \atabl{rewards_accuracy.details}.

Surprisingly, the linear classifier performs well and does not appear to overfit on our relatively small training set. Although the feature selection algorithm is rather close to the linear classifier compared to the baseline, using feature selection to avoid overfitting is not necessary. However the idea that a small subset of features (32 in this case) can lead to reasonable classification accuracy is verified and an important piece of information for drastically reducing the search space for future work in unsupervised steps discovery. Additionally, we show in \fig{features_graph} that the feature selection approach works well when the number of features $n$ is in the region $[32, 64]$ but collapses to $0\%$ accuracy when $n > 8192$.

\subsection{Real-World Robotic Door Opening}
\label{sec:door}

In this section, we aim to answer the question of whether our previously visualized reward function can be used to learn a real-world robotic motion skill. We experiment on a door opening skill, where we adapt a demonstrated door opening to a novel configuration, such as different position or orientation of the door. Following the experimental protocol in prior work \citep{chebotar2016path}, we adapt an imperfect kinesthetic demonstration which we ensure succeeds occasionally (about 10\% of the time). These demonstrations consist only of robot poses, and do not include images. We then use a variety of different video demonstrations, which contain images but not robot poses, to learn the reward function. These videos include demonstrations with other doors, and even demonstrations provided by a human using their own body, rather than through kinesthetic teaching with the robot. Across all experiments, we use a total of 11 videos. We provide videos of experiments in \footnote{\href{https://sermanet.github.io/rewards/}{sermanet.github.io/rewards}}.

\begin{figure}[h]
\begin{center}
\includegraphics[width=.6\linewidth]{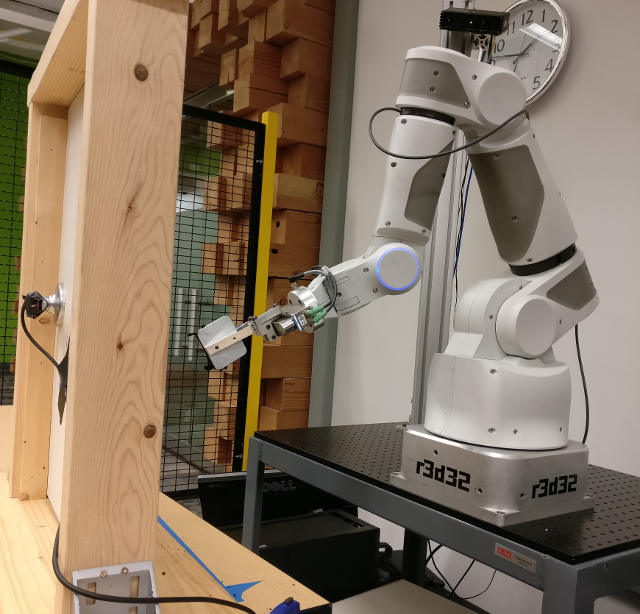}
\end{center}
\caption{\small{{\bf Robot arm setup.} Note that our method does not make use of the sensor on the back handle of the door, but it is used in our comparison to train a baseline method with the ground truth reward.}}
\label{setup}
\end{figure}

Figure~\ref{setup} shows the experimental setup. We use a 7-DoF robotic arm with a two-finger gripper, and a camera placed above the shoulder, which provides monocular RGB images.
For our baseline PI$^2$ policy, we closely follow the setup of \cite{chebotar2016path} which uses an IMU sensor in the door handle to provide both a cost and feedback as
part of the state of the controller. In contrast, in our approach we remove this sensor both from the state representation provided to PI$^2$
and in our reward replace the target IMU state with the output of a deep neural network.

\begin{figure*}[htb]
\begin{center}

\adjustbox{minipage=1.3em,valign=t}{\subcaption{}\label{fig.r3human.human_08}}
\begin{subfigure}[t]{0.9\linewidth}
  \includegraphics[width=1\linewidth]{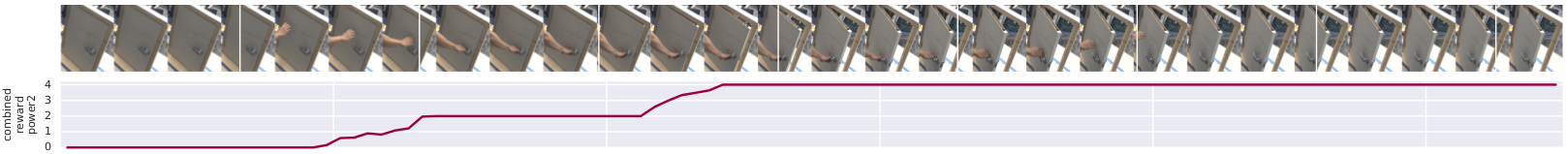}
\end{subfigure}

\adjustbox{minipage=1.3em,valign=t}{\subcaption{}\label{fig.r3human.human-miss_10}}
\begin{subfigure}[t]{0.9\linewidth}
  \includegraphics[width=1\linewidth]{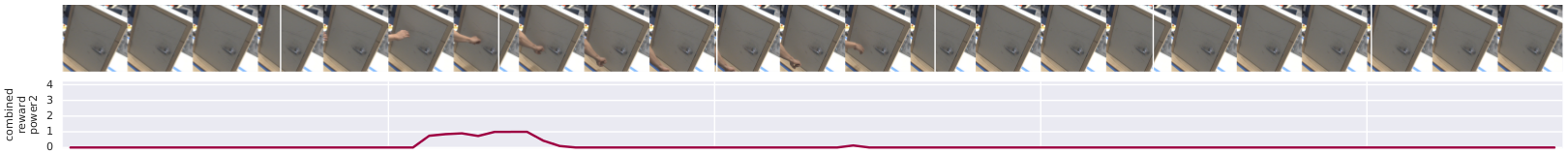}
\end{subfigure}

\adjustbox{minipage=1.3em,valign=t}{\subcaption{}\label{fig.r3human.17_01}}
\begin{subfigure}[t]{0.9\linewidth}
  \includegraphics[width=1\linewidth]{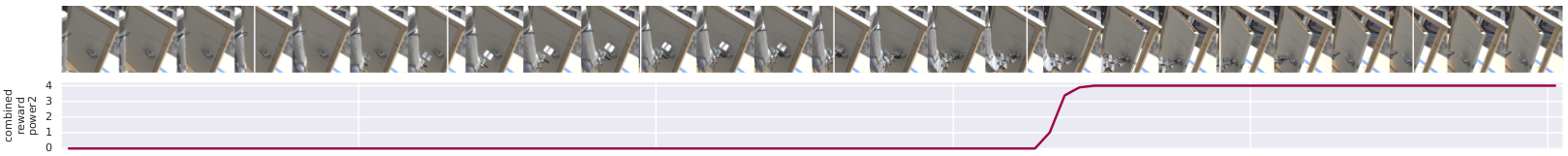}
\end{subfigure}

\adjustbox{minipage=1.3em,valign=t}{\subcaption{}\label{fig.r3human.blackdoor_00}}
\begin{subfigure}[t]{0.9\linewidth}
  \includegraphics[width=1\linewidth]{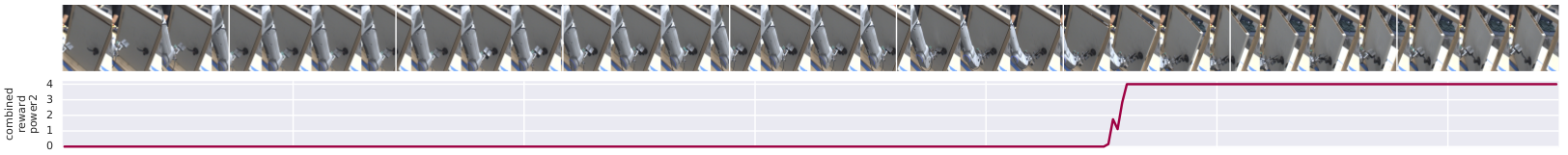}
\end{subfigure}

\adjustbox{minipage=1.3em,valign=t}{\subcaption{}\label{fig.r3human.bluedoor_00}}
\begin{subfigure}[t]{0.9\linewidth}
  \includegraphics[width=1\linewidth]{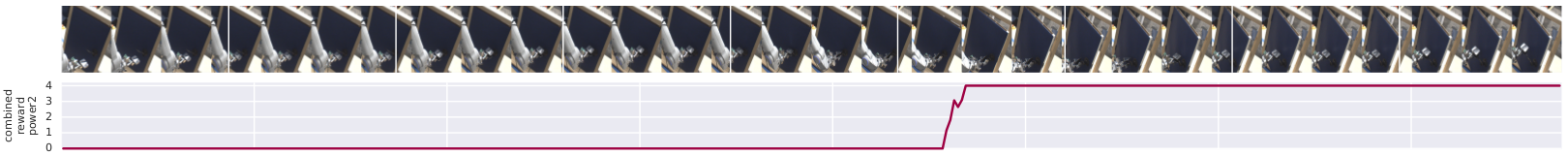}
\end{subfigure}

\end{center}
\caption{{\bf Rewards from human demonstration only.}
Here we show the rewards produced when trained on humans only (see \afig{r3human.training}).
In \ref{fig.r3human.human_08}, we show the reward on a human test video.
In \ref{fig.r3human.human-miss_10}, we show what the reward produces when the human hands misses opening the door.
In \ref{fig.r3human.17_01}, we show the reward successfully saturates when the robot opens the door even though it has not seen a robot arm before. Similarly in \ref{fig.r3human.blackdoor_00} and \ref{fig.r3human.bluedoor_00} we show it still works with some amount of variation of the door which was not seen during training (white door and black handle, blue door, rotations of the door).}
\label{fig.r3human.rewards}
\end{figure*}

\subsubsection{Data}

We experiment with a range of different demonstrations from which we derive our reward function, varying both the source demo (human vs robotic), the number of subgoals we extract, and the appearance of the door. We record monocular RGB images on a camera placed above the shoulder of the arm. The door is cropped from the images, and then the resulting image is re-sized such that the shortest side is 299 dimensional with preserved aspect ratio. The input into our convolutional feature extractor  \cite{DBLP:journals/corr/SzegedyVISW15} is the 299x299 center crop. 

\subsubsection{Qualitative Analysis}
\label{sec.real.qualitative}

\begin{figure}[htb!]
\begin{center}
\includegraphics[width=1\linewidth]{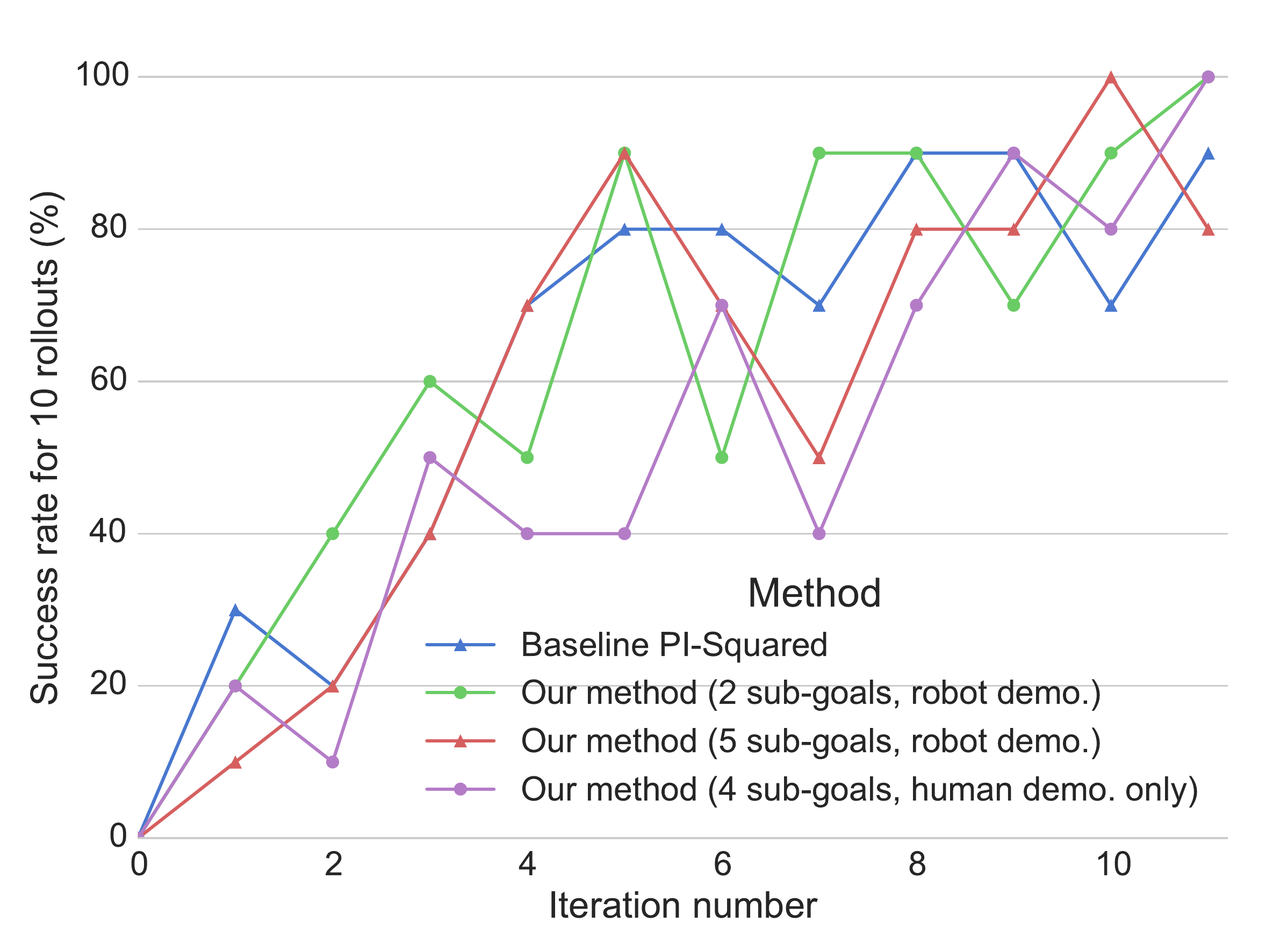}
\end{center}
\caption{{\bf Door opening success rate at each iteration of learning on the real robot.} The PI$^2$ baseline method uses a ground truth reward function obtained by instrumenting the door. Note that rewards learned by our method, even from videos of humans or different doors, learn comparably or faster when compared to the ground truth reward.}
\label{fig.quant.real}
\end{figure}

We evaluate our reward functions qualitatively by plotting our
perceptual reward functions below the demonstrations with a variety of 
door types and demonstrators (e.g robot or human). As can be seen in \fig{r3human.rewards} and in real experiments \fig{quant.real}, we show that the reward functions are useful to a robotic arm while only showing human demonstrations as depicted in \afig{r3human.training}. Moreover we exhibit robustness variations in appearance.

\subsubsection{Quantitative Analysis}

In comparing the success rate of visual reward versus a baseline PI$^2$ method that use s
the ground truth reward function obtained by instrumenting the door with an IMU. We run PI$^2$ for 11 iterations with 10 sampled trajectories at each iteration. As can be seen in \fig{quant.real}, we obtain similar convergence speeds to our baseline model, with our method also able to open the door consistently. Since our local policy is able to 
obtain high reward candidate trajectories, this is strong evidence that a perceptual reward could be used to train a global policy in same manner as \cite{chebotar2016path}. 

\section{Conclusion}

In this paper, we present a method for automatically identifying important intermediate goal given a few visual demonstrations of a task. By leveraging the general features learned from pre-trained deep models, we propose a method for rapidly learning an incremental reward function from human demonstrations which we successfully demonstrate on a real robotic learning task.

We show that pre-trained models are general enough to be used without retraining. We also show there exists a small subset of pre-trained features that are highly discriminative even for previously unseen scenes and which can be used to reduce the search space for future work in unsupervised subgoal discovery.

In this work, we studied imitation in a setting where the viewpoint of the robot/demonstrator is fixed. An interesting avenue of future
work would be to analyze the impact of changes in viewpoint. The ability to learn from a learn from broad diverse sets of experience also ties
into the goal of lifelong robotic learning. Continuous learning using unsupervised rewards promises to substantially increase the variety and diversity of experiences, resulting in more robust and general robotic skills. 

\subsubsection*{Acknowledgments}
We thank Vincent Vanhoucke for helpful discussions and feedback, Mrinal Kalakrishnan and Ali Yahya for indispensable guidance throughout this project and Yevgen Chebotar for his PI$^2$ implementation. We also thank the anonymous reviewers for their feedback and constructive comments.

%% file: rss_appendix.tex
\appendix

\section{Algorithms Details}

\subsection{Binary Segmentation Algorithm}
\label{appendix.binary_similarity}

\begin{algorithm}[htb]
  \begin{algorithmic}
\small{
  \Function{BinarySplit}{$video$, $start$, $end$, $n$, $min\_size$, $prev\_std=[]$}
  \If{$n = 1$}
  \Return $[], []$
  \EndIf
  
  \State $splits0, std0 \leftarrow \Call{Split}{video, start, end, 2, min\_size}$
  \If{$n = 2$}
  \Return $splits0, std0$
  \EndIf
  
  \State $splits1, std1 \leftarrow \textsc{BinarySplit}(\parbox[t]{.4\linewidth}{$video$, $start$, $splits0[0]$, $\Call{Ceil}{n / 2}$, $min\_size$)}$
  
  \State $splits2, std2 \leftarrow \textsc{BinarySplit}(\parbox[t]{.4\linewidth}{$video$, $splits0[0] + 1$, $end$, $\Call{Floor}{n / 2}$, $min\_size$)}$

  \State $all\_splits = []$
  \State $all\_std = []$
  
  \If{$splits1 \neq []$}
    \State $\Call{Join}{all\_splits, splits1}$
    \State $\Call{Join}{all\_std, std1}$
  \Else
    \State $\Call{Join}{all\_std, std0[0]}$
  \EndIf

  \If{$splits0 \neq []$}
    \State $\Call{Join}{all\_splits, splits0[0]}$
  \EndIf

  \If{$splits2 \neq []$}
    \State $\Call{Join}{all\_splits, splits2}$
    \State $\Call{Join}{all\_std, std2}$
  \Else
    \State $\Call{Join}{all\_std, std0[1]}$
  \EndIf
  \State \Return $all\_splits, all\_std$
\EndFunction
\caption{Greedy and binary algorithm similar to and utilizing Algorithm~\ref{algo.similarity}, where
$AverageStd()$ is a function that computes the average standard deviation over a set of frames or over a set of values, $Join()$ is a function that joins values or lists together into a single list, $n$ is the number of splits desired and $min\_size$ is the minimum size of a split.}
\label{algo.binary_similarity}
}
\end{algorithmic}
\end{algorithm}

\begin{table*}[htb]
\label{table.unsup_steps_accuracy.details}
\small
\caption{{\bf Per-step details of unsupervised steps discovery accuracy} (Jaccard overlap on training sets) versus the ordered random steps baseline.}\label{appendix.table.split_accuracy}
\begin{center}
\begin{tabular}{c|c|c|c|c|c|c|c|c}
\multicolumn{1}{c|}{\bf dataset} & \multicolumn{1}{c|}{\bf method} & \multicolumn{3}{c|}{\bf 2 steps}  & \multicolumn{4}{c}{\bf 3 steps}\\
(training)& & step 1 & step 2 & average & step 1 & step 2 & step 3 & average\\
\hline
door & ordered random steps           & 59.4\% &     45.6\% &     52.5\% &     48.0\% &     58.1\% &     60.1\% &     55.4\% \\
     & unsupervised steps & \bf 84.0\% & \bf 68.1\% & \bf 76.1\% & \bf 57.6\% & \bf 75.1\% & \bf 68.1\% & \bf 66.9\% \\
 \hline
pouring & ordered random steps &           65.2\% &     66.6\% &     65.9\% &     46.2\% &     46.3\% & \bf 66.3\% &     52.9\% \\
        & unsupervised steps & \bf 92.3\% & \bf 90.5\% & \bf 91.6\% & \bf 79.7\% & \bf 48.0\% & 48.6\% &     \bf 58.8\% \\
\end{tabular}
\end{center}
\end{table*}

\subsection{Combining Intermediate Rewards}
\label{appendix.combining}

\begin{figure*}[htb]
\begin{center}
\includegraphics[width=.9\linewidth]{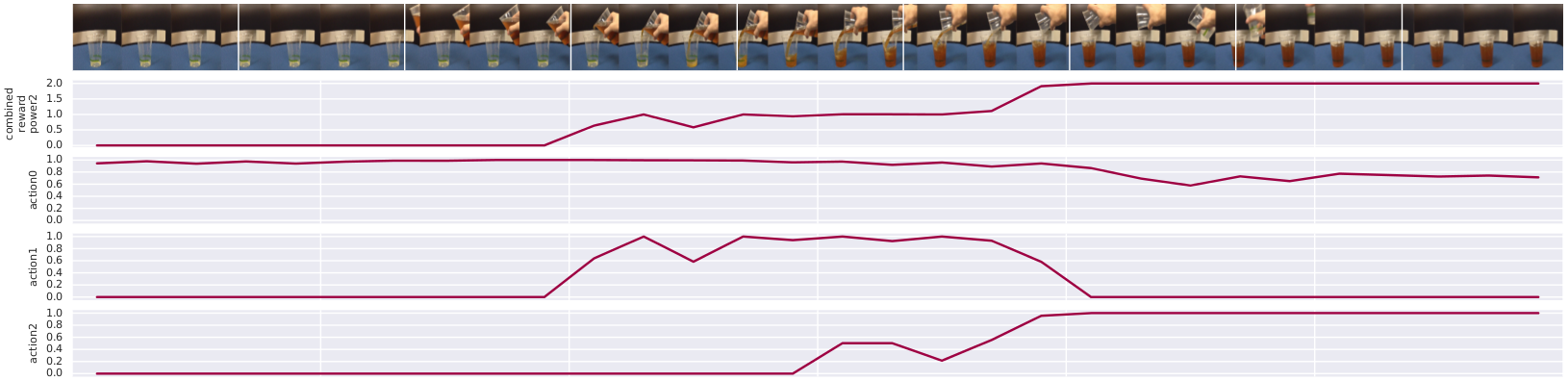}
\end{center}
\caption{{\bf Combining intermediate rewards into a single reward function.} From top to bottom, we show the combined reward function (with range [0,2]) followed by the reward function of each individual steps (with range [0,1]). The first step reward corresponds to the initial resting state of the demonstration and is ignored in the reward function. The second step corresponds to the pouring action and the third step corresponds to the glass full of liquid. }
\label{fig.combination.pouring}
\end{figure*}

From the two previous sections, we obtain one reward function per intermediate step discovered by the unsupervised algorithm. These need to be combined so that the RL algorithm uses a single reward function which partially rewards intermediate steps but emphasizes later rewards. The initial step is ignored as it is assumed to be the initial starting state in the demonstrations. We opt for the maximum range of each reward to be twice the maximum range of its preceding reward, summing them as follow:
\begin{equation}
    R(a) = \sum_{i=2}^{n} R_i(a) * 2^{(i - 1)}
\end{equation}
where $n$ is the number of intermediate rewards detected and $a$ an activations vector. An example of this combination is shown in \fig{combination.pouring}.

\subsection{Feature Selection Algorithm}
\label{appendix.selection}

Here we describe the feature selection algorithm we use to investigate the presence of a small subset of discriminative features in mid to high level layers of a pre-trained deep network.
To select the most discriminative features, we use a simple scoring heuristic. Each feature $i$ is first normalized by subtracting the mean and dividing by the standard deviation of all training sequences. We then rank them for each sub-goal according to their distance $z_i$ to the average statistics of the sets of positive and negative frames for a given goal:
\begin{equation}
  z_i = \alpha \abs{\mu_i^{+} - \mu_i^{-}} - (\sigma_i^{+} + \sigma_{i}^{-}),
\end{equation}
where $\mu_i^{+}$ and $\sigma_i^{+}$ are the mean and standard deviation of all ``positive" frames and the $\mu_i^{-}$ and $\sigma_i^{-}$ of all ``negative" frames (the frames that do not contain the sub-goal). Only the top-$M$ features are retained to form the reward function $R_g()$ for the sub-goal $g$, which is given by the log-probability of an independent Gaussian distribution over the relevant features:
\begin{equation}
  R_g(s_t) = \frac{1}{n}\sum_{j}^{M} \frac{(s_{{i_j}t} - \mu_{{i_j}t}^{+})^2}{{\sigma_{{i_j}t}^{+}}^2},
\end{equation}
where $i_j$ indexes the top-$M$ selected features. We empirically choose $\alpha = 5.0$ and $M = 32$ for our subsequent experiments. At test time, we do not know when the system transitions from one goal to another, so instead of time-indexing the goals, we instead combine all of the goals into a single time-invariant reward function, where later steps yield higher reward than earlier steps, as described in Appendix~\ref{appendix.combining}.

\begin{table*}[htb]
\caption{{\bf Reward functions accuracy} by steps (Jaccard overlap on test sets).}
\label{appendix.table.rewards_accuracy.details}
\small
\begin{center}
\begin{tabular}{c|c|c|c|c|c|c|c|c|}
\multicolumn{1}{c|}{\bf dataset} & \multicolumn{1}{c|}{\bf method} & \multicolumn{4}{c|}{\bf steps}  \\
(testing)& & step 1 & step 2 & step 3 & average \\
\hline
door & random baseline           & $40.8\% \pm 1.0$ & $26.3\% \pm 4.1 $ & - & $33.6\% \pm 1.6$ \\
2-steps     & feature selection          & $ 85.1\% \pm 0.0$ & $ 59.7\% \pm 0.0 $ & - & $ 72.4\% \pm 0.0$  \\
     & linear classifier  & $79.7\% \pm 6.0$ & $70.4\% \pm 5.0 $& - & ${\bf 75.0\% \pm 5.5} $  \\
\hline
door & random baseline           &  $20.8\% \pm 1.1$ & $31.8\% \pm 1.6$ & $23.8\% \pm 2.3 $ & $25.5\% \pm 1.6$ \\
3-steps     & feature selection          & $ 56.9\% \pm 0.0 $ & $ 47.7\% \pm 0.0$ & $ 54.1\% \pm 0.0$ & $ 52.9\% \pm 0.0$ \\
     & linear classifier  &  $46\% 6.9\pm $ & $47.5\% \pm 4.2 $ & $ 67.2\% \pm 3.3 $ & ${\bf 53.6\% \pm 4.7 }$ \\
 \hline
pouring & random baseline   &     $ 39.2\% \pm 2.9 $ & $22.9\% \pm 3.9 $ & - & $31.1\% \pm 3.4 $ \\
2-steps        & feature selection & $ 76.2\% \pm 0.0$ & $54.6\% \pm 0.0$ & - & $65.4\% \pm 0.0$ \\
        & linear classifier & $78.2\% \pm 2.4$ & $ 60.2\% \pm 1.7$ & - & $ {\bf 69.2\% \pm 2.0}$ \\
 \hline
pouring & random baseline   &  $22.5\% \pm 0.6 $ & $38.8\% \pm 0.8 $ & $ 13.9\% \pm 0.1 $ & $ 25.1\% \pm 0.1 $\\
3-steps        & feature selection & $32.9\% \pm 0.0$ & $55.2\% \pm 0.0$ & $32.2\% \pm 0.0$ & $40.0\% \pm 0.0$ \\
        & linear classifier &  $72.5\% \pm 10.5$ & $37.2\% \pm 11.0 $ & $ 39.1\% \pm 6.8$ & ${\bf 49.6\% \pm 8.0}$\\
\end{tabular}
\end{center}
\end{table*}

\subsection{PI$^2$ Reinforcement Learning Algorithm}
\label{appendix.rl_details}

We chose the PI$^2$ reinforcement learning algorithm \citep{theodorou2010generalized} for our experiments, with the particular implementation of the method based on a recently proposed deep reinforcement learning variant \citep{chebotar2016path}. Since our aim is mainly to validate that our learned reward functions capture the goals of the task well enough for learning, we employ a relatively simple linear-Gaussian parameterization of the policy, which corresponds to a sequence of open-loop torque commands with fixed linear feedback to correct for perturbations, as in the work of \cite{chebotar2016path}. This policy has the form $\pi(\mathbf{u}_t|\mathbf{x}_t) = \mathcal{N}(\mathbf{K}_t \mathbf{x}_t + \mathbf{k}_t, \Sigma_t)$, where $\mathbf{K}_t$ is a fixed stabilizing feedback matrix, and $\mathbf{k}_t$ is a learned control. In this case, the state $\mathbf{x}_t$ corresponds to the joint angles and angular velocities of a robot, and $\mathbf{u}_t$ corresponds to the joint torques. Since the reward function is evaluated from camera images, we assume that the image is a (potentially stochastic) consequence of the robot's state, so that we can evaluate the state reward $r(\mathbf{x}_t)$ by taking the image $\mathbf{I}_t$ observed at time $t$, and computing the corresponding activations $a_t$. Overloading the notation, we can write $a_t = f(\mathbf{I}_t(\mathbf{x}_t))$, where $f$ is the network we use for visual features. Then, we have $r(\mathbf{x}_t) = R(f(\mathbf{I}_t(\mathbf{x}_t)))$.

The PI$^2$ algorithm is an episodic policy improvement algorithm that uses the reward $r(\mathbf{x}_t)$ to iteratively improve the policy. The trust-region variant of PI$^2$ that we use \cite{chebotar2016path}, which is also similar to the REPS algorithm~\citep{reps}, updates the policy at iteration $n$ by sampling from the time-varying linear-Gaussian policy $\pi(\mathbf{u}_t | \mathbf{x}_t)$ to obtain samples $\{(\mathbf{x}_t^{(i)},\mathbf{u}_t^{(i)})\}$, and updating the controls $\mathbf{k}_t$ at each time step according to
\[
\mathbf{k}_t \leftarrow \left[ \sum_i \mathbf{u}_t^{(i)} \exp\left(\beta_t\sum_{t' = t}^T r(\mathbf{x}_{t'}^{(i)}) \right)\right] / \left[\sum_i \exp\left(\beta_t\sum_{t' = t}^T r(\mathbf{x}_{t'}^{(i)}) \right)\right],
\]%
where the temperature $\beta_t$ is chosen to bound the KL-divergence between the new policy $\pi(\mathbf{u}_t|\mathbf{x}_t)$ and the previous policy $\bar{\pi}(\mathbf{u}_t|\mathbf{x}_t)$, such that $D_\text{KL}(\pi(\mathbf{u}_t|\mathbf{x}_t) \| \bar{\pi}(\mathbf{u}_t|\mathbf{x}_t)) \leq \epsilon$ for a step size epsilon. Further details and a complete derivation are provided in prior work \cite{theodorou2010generalized,reps,chebotar2016path}.

The PI$^2$ algorithm is a local policy search method that performs best when provided with demonstrations to bootstrap the policy. In our experiments, we use this method together with our learned reward functions to learn a door opening skill with a real physical robot, as discussed in Section~\ref{sec:door}. Demonstration are provided with kinesthetic teaching, which results in a sequence of reference steps $\hat{\mathbf{x}}_t$, and initial controls $\mathbf{k}_t$ are given by $\mathbf{k}_t = -\mathbf{K}_t \hat{\mathbf{x}}_t$, such that the mean of the initial controller is $\mathbf{K}_t (\mathbf{x}_t - \hat{\mathbf{x}}_t)$, corresponding to a trajectory-following initialization. This initial controller is rarely successful consistently, but the occasional successes it achieves provide a learning signal to the algorithm. The use of demonstrations enables PI$^2$ to be used to quickly and efficiently learn complex robotic manipulation skills.

Although this particular RL algorithm requires demonstrations to begin learning, it can still provide a useful starting point for real-world learning with a real robotic system. As shown by previous work\cite{chebotar2016path}, the initial set of demonstrations can be expanded into a generalizable policy by iteratively ``growing'' the effective region where the policy succeeds. For example, if the robot is provided with a demonstration of opening a door in one position, additional learning can expand the policy to succeed in nearby positions, and the application of a suitable curriculum can grow the region of door poses in which the policy succeeds progressively. However, as with all RL algorithms, this process requires knowledge of the reward function. Using the method described in this paper, we can learn such a reward function from either the initial demonstrations or even from other demonstration videos provided by a human. Armed with this learned reward function, the robot could continue to improve its policy through real-world experience, iteratively increasing its region of competence through lifelong learning.

\section{Additional Qualitative Results}

\begin{figure*}[htb]
\begin{center}
\includegraphics[width=1\linewidth]{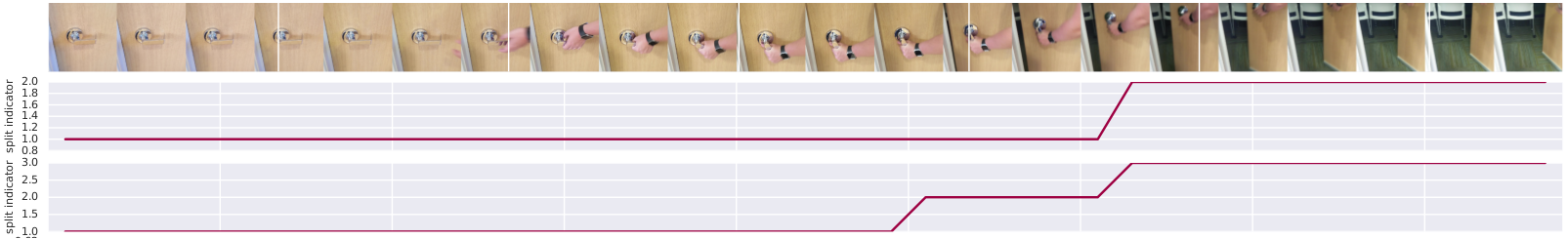}
\includegraphics[width=1\linewidth]{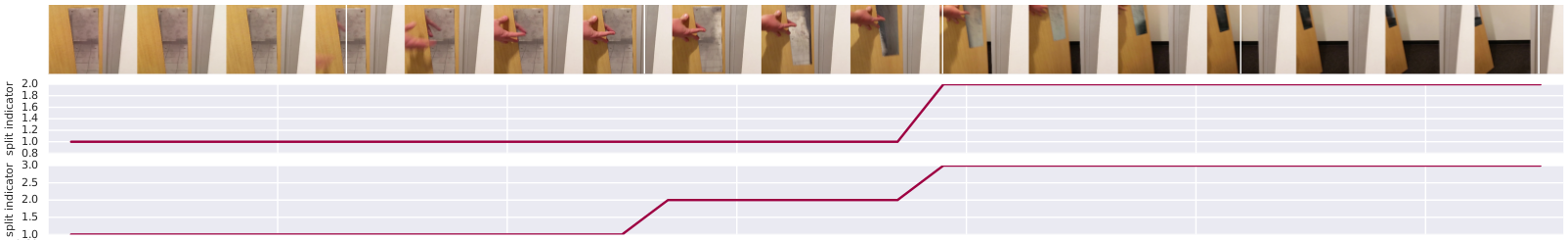}
\includegraphics[width=1\linewidth]{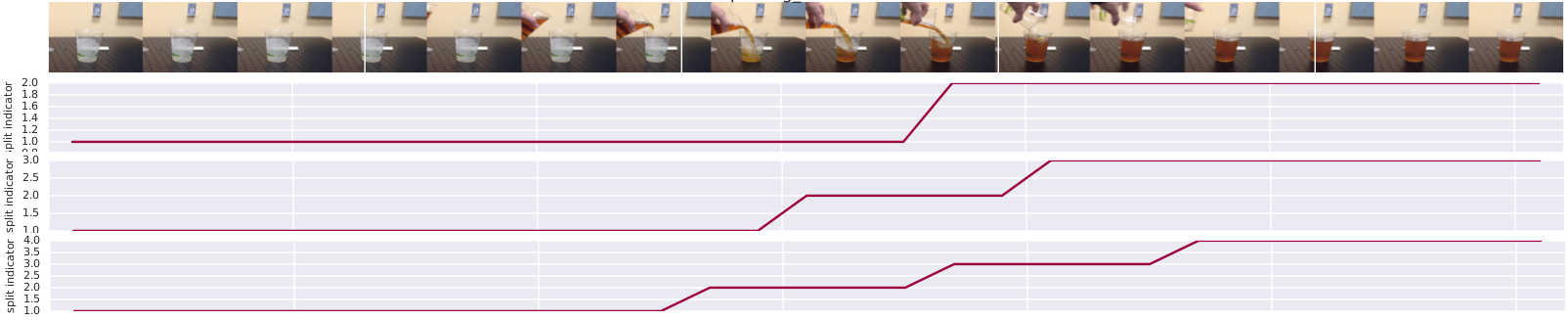}
\includegraphics[width=1\linewidth]{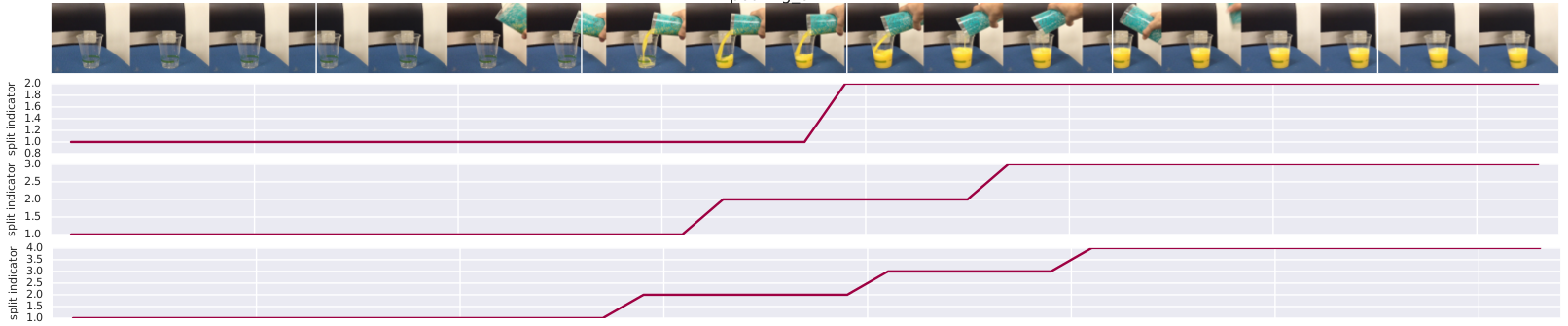}
\end{center}
\caption{{\bf Qualitative examples of unsupervised discovery of steps for door and pouring tasks} in training videos. For each video, we show the detected splits when splitting in 2, 3 or 4 segments. Each segment is delimited by a different value on the vertical axis of the curves.}
\label{fig.splits}
\end{figure*}

\begin{figure*}[htb]
\begin{center}

\adjustbox{minipage=1.3em,valign=t}{\subcaption{}\label{fig.X_11}}
\begin{subfigure}[t]{1\textwidth}
  \includegraphics[width=1\linewidth]{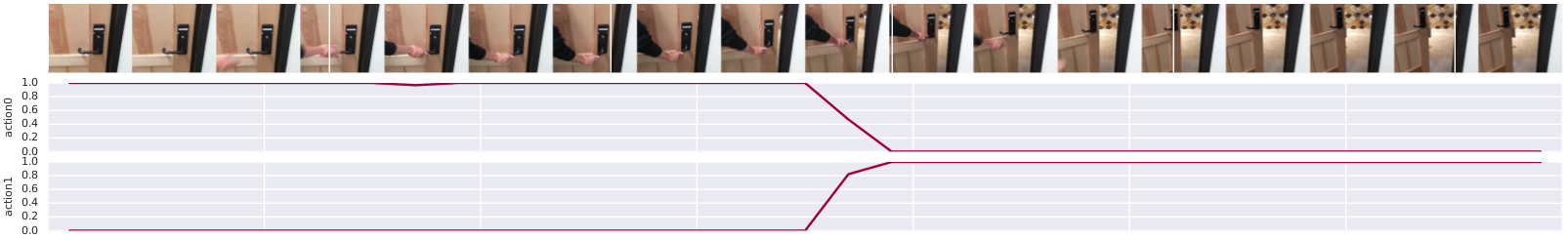}
\end{subfigure}

\adjustbox{minipage=1.3em,valign=t}{\subcaption{}\label{fig.X_11_1}}
\begin{subfigure}[t]{1\textwidth}
  \includegraphics[width=1\linewidth]{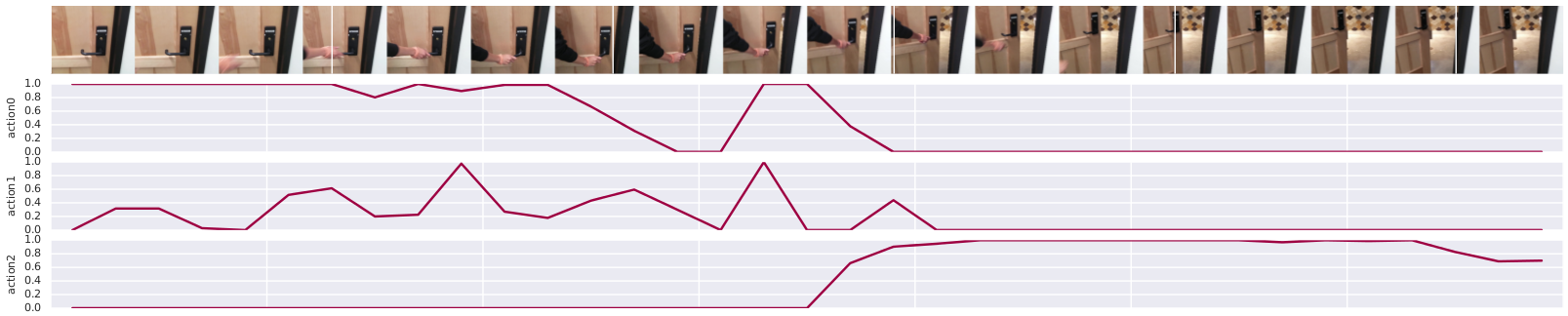}
\end{subfigure}

\adjustbox{minipage=1.3em,valign=t}{\subcaption{}\label{fig.hobbit_7}}
\begin{subfigure}[t]{1\textwidth}
  \includegraphics[width=1\linewidth]{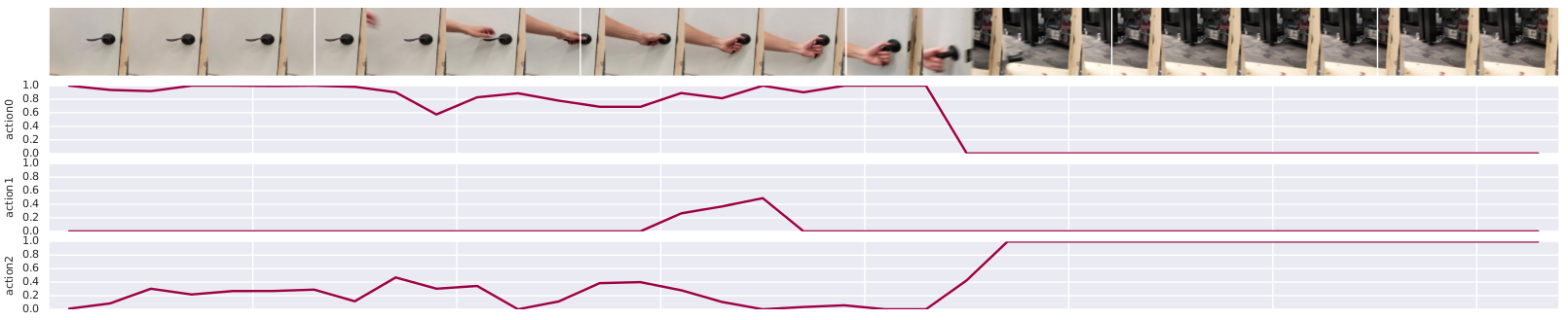}
\end{subfigure}

\adjustbox{minipage=1.3em,valign=t}{\subcaption{}\label{fig.1965_13}}
\begin{subfigure}[t]{1\textwidth}
  \includegraphics[width=1\linewidth]{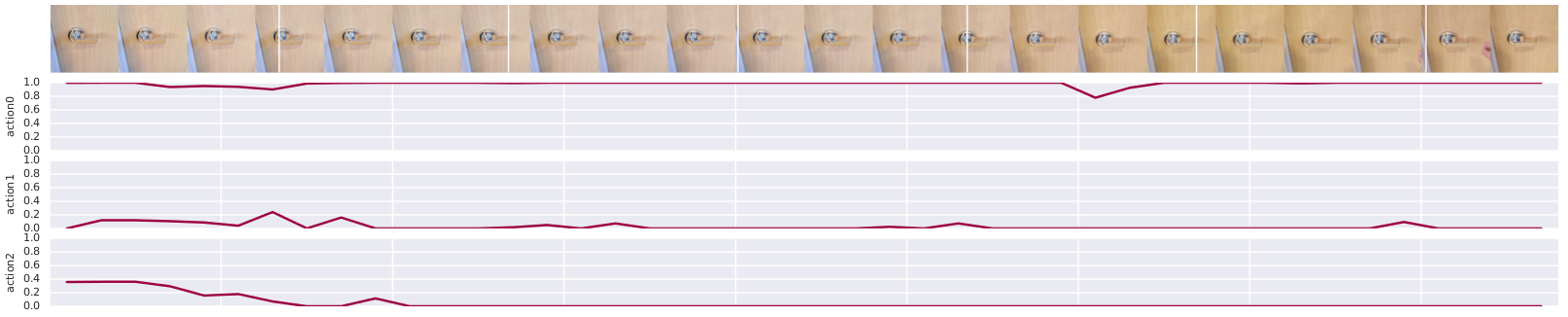}
\end{subfigure}

\adjustbox{minipage=1.3em,valign=t}{\subcaption{}\label{fig.1965_0}}
\begin{subfigure}[t]{1\textwidth}
  \includegraphics[width=1\linewidth]{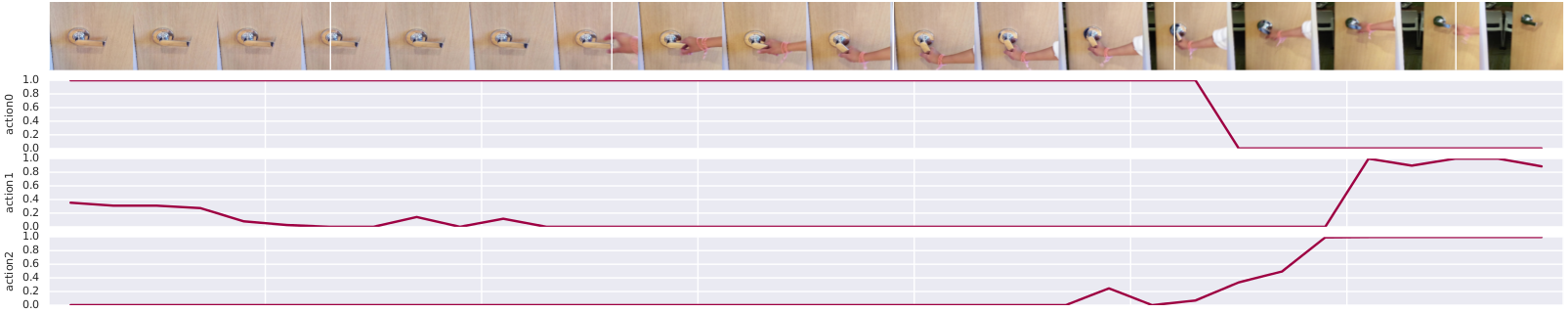}
\end{subfigure}

\end{center}
\caption{{\bf Qualitative examples of reward functions for the door task} in testing videos. These plots show the individual sub-goal rewards for either 2 or 3 goals splits. The "open" or "closed" door reward functions are firing quite reliably in all plots, the "hand on handle" step however can be a weaker and noisier signal as seen in \ref{fig.X_11_1} and \ref{fig.hobbit_7}, or incorrect as shown in \ref{fig.1965_0}.
\ref{fig.1965_13} demonstrates how a "missed" action is correctly recognized. 
}
\label{fig.qual.door}
\end{figure*}

\begin{figure*}[htb]
\begin{center}

\adjustbox{minipage=1.3em,valign=t}{\subcaption{}\label{fig.pouring_007}}
\begin{subfigure}[t]{.9\textwidth}
  \includegraphics[width=1\linewidth]{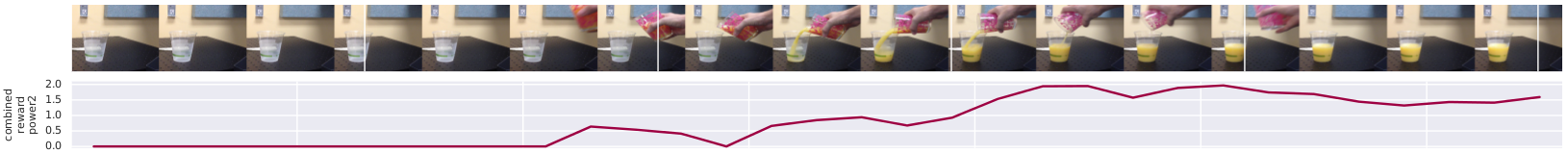}
\end{subfigure}

\adjustbox{minipage=1.3em,valign=t}{\subcaption{}\label{fig.pouring_012}}
\begin{subfigure}[t]{.9\textwidth}
  \includegraphics[width=1\linewidth]{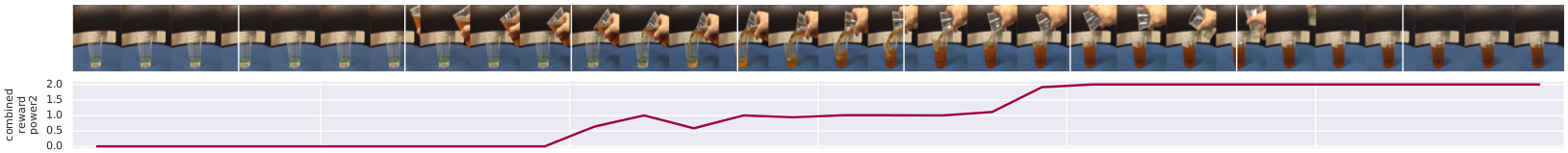}
\end{subfigure}

\adjustbox{minipage=1.3em,valign=t}{\subcaption{}\label{fig.pouring_016}}
\begin{subfigure}[t]{.9\textwidth}
  \includegraphics[width=1\linewidth]{{\figdircombined}/pouring_016.png}
\end{subfigure}

\adjustbox{minipage=1.3em,valign=t}{\subcaption{}\label{fig.pouring_001}}
\begin{subfigure}[t]{.9\textwidth}
  \includegraphics[width=1\linewidth]{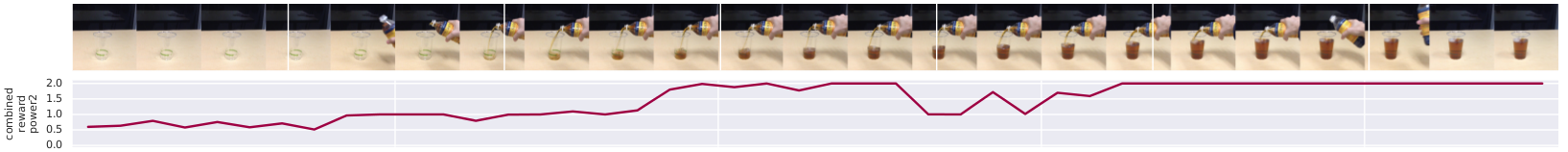}
\end{subfigure}

\adjustbox{minipage=1.3em,valign=t}{\subcaption{}\label{fig.pouring_004}}
\begin{subfigure}[t]{.9\textwidth}
  \includegraphics[width=1\linewidth]{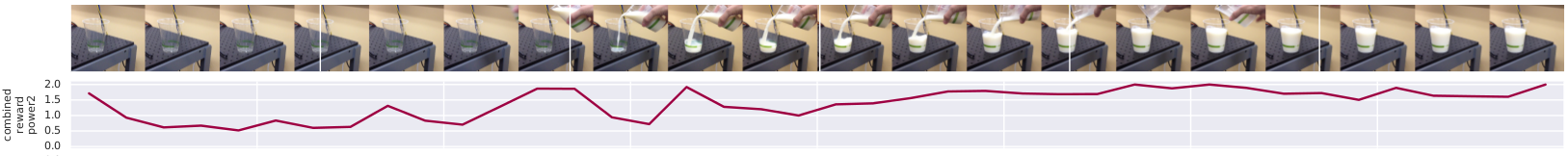}
\end{subfigure}

\adjustbox{minipage=1.3em,valign=t}{\subcaption{}\label{fig.pouring_009}}
\begin{subfigure}[t]{.9\textwidth}
  \includegraphics[width=1\linewidth]{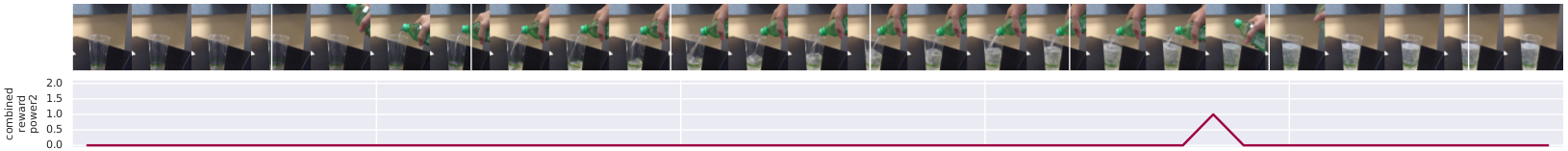}
\end{subfigure}

\adjustbox{minipage=1.3em,valign=t}{\subcaption{}\label{fig.pouring-miss_001}}
\begin{subfigure}[t]{.9\textwidth}
  \includegraphics[width=1\linewidth]{{\figdircombined}/pouring-miss_001.png}
\end{subfigure}

\adjustbox{minipage=1.3em,valign=t}{\subcaption{}\label{fig.pouring-miss_002}}
\begin{subfigure}[t]{.9\textwidth}
  \includegraphics[width=1\linewidth]{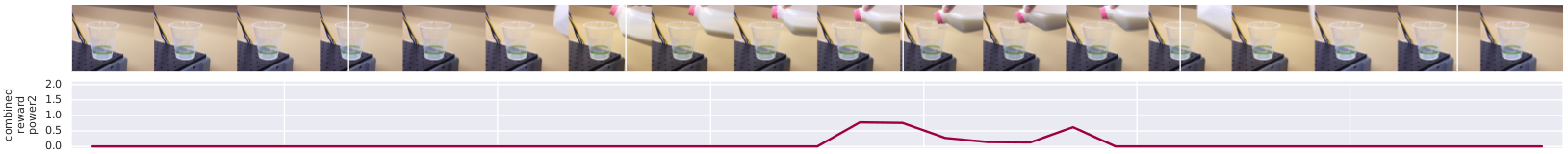}
\end{subfigure}

\adjustbox{minipage=1.3em,valign=t}{\subcaption{}\label{fig.pouring-miss_003}}
\begin{subfigure}[t]{.9\textwidth}
  \includegraphics[width=1\linewidth]{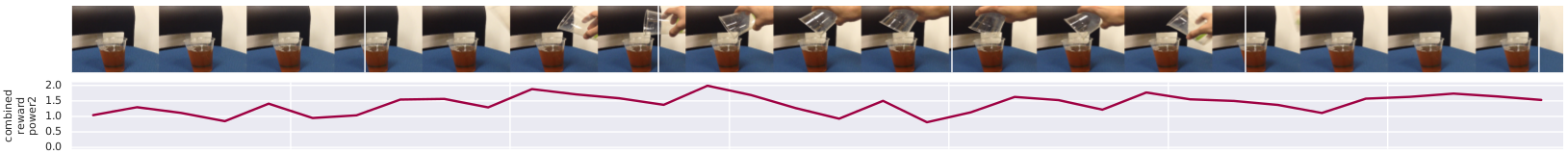}
\end{subfigure}

\adjustbox{minipage=1.3em,valign=t}{\subcaption{}\label{fig.pouring-miss_004}}
\begin{subfigure}[t]{.9\textwidth}
  \includegraphics[width=1\linewidth]{{\figdircombined}/pouring-miss_004.png}
\end{subfigure}

\end{center}
\caption{{\bf Entire testing set of "pouring" reward functions.}
This testing set is designed to be more challenging than the training set by including ambiguous cases such as pouring into an already full glass (\ref{fig.pouring-miss_003} and \ref{fig.pouring-miss_004}) or pouring with a closed bottle (\ref{fig.pouring-miss_001} and \ref{fig.pouring-miss_002}). Despite the ambiguous inputs, the reward functions do produce reasonably low or high reward based on how full the glass is. \ref{fig.pouring_007}, \ref{fig.pouring_012}, \ref{fig.pouring_012} and \ref{fig.pouring_001} are not strictly monotonically increasing but do overall demonstrate a reasonable progression as the pouring is executed to a saturated maximum reward when the glass is full. \ref{fig.pouring_004} also correctly trends upwards but starts with a high reward with an empty glass.
\ref{fig.pouring_009} is a failure case where the somewhat transparent liquid is not detected.}
\label{fig.qual.rewards}
\end{figure*}

\begin{figure*}[h]
\begin{center}
\includegraphics[width=1\linewidth]{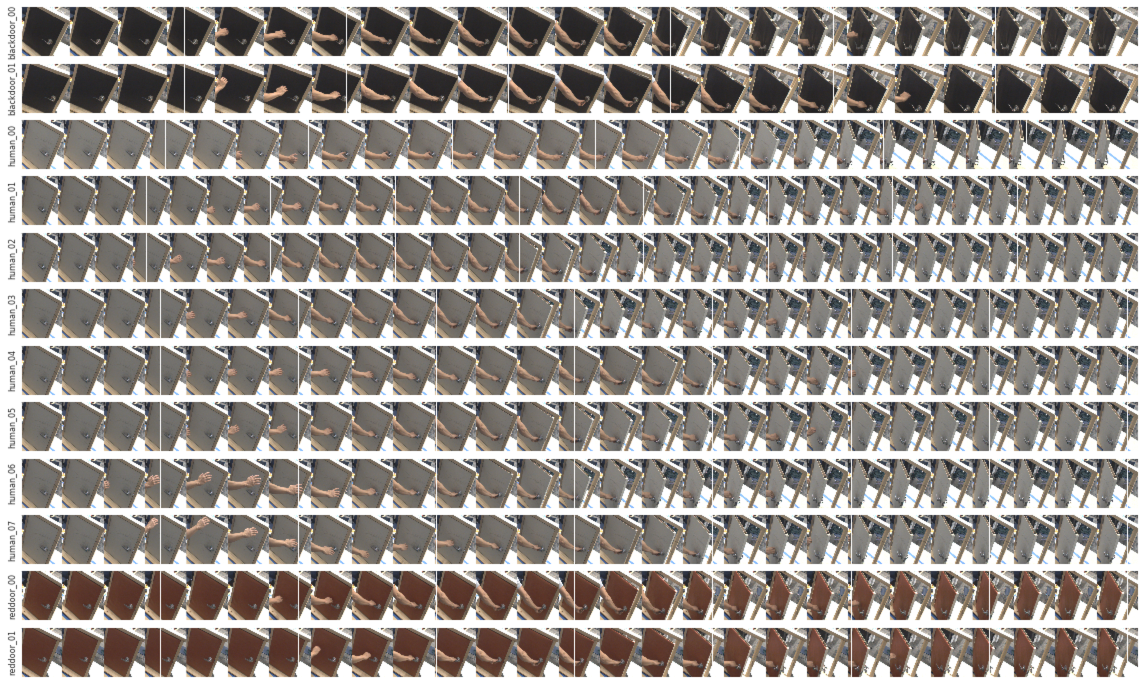}
\end{center}
\caption{{\bf Entire training set of human demonstrations.}}
\label{fig.r3human.training}
\end{figure*}